\documentclass[runningheads]{llncs}

\usepackage{graphicx}
\usepackage{amsmath}
\usepackage{amssymb}
\usepackage{algorithm}
\usepackage{algpseudocode}

\usepackage{booktabs}
\usepackage{multirow}
\usepackage{hyperref}
\usepackage[noblocks]{authblk}

\begin{document}

\title{Sequential Counterfactual Inference for Temporal Clinical Data: Addressing the Time Traveler Dilemma}

\author{Jingya Cheng\inst{1, *} \and
Alaleh Azhir\inst{1, *} \and
Jiazi Tian\inst{1} \and Hossein Estiri\inst{1, +}}
\authorrunning{Cheng et al.}
%
\institute{Massachusetts General Hospital, Boston, MA, USA
\email{hestiri@mgh.harvard.edu}\\ \and
*Equal contribution (co–first authors)\\ \and
+Corresponding author\\
}

\maketitle

\begin{abstract}
Counterfactual inference enables clinicians to ask ``what if'' questions about patient outcomes, but standard methods assume feature independence and simultaneous modifiability---assumptions violated by longitudinal clinical data. We introduce the \textbf{Sequential Counterfactual Framework}, which respects temporal dependencies in electronic health records by distinguishing immutable features (chronic diagnoses) from controllable features (lab values) and modeling how interventions propagate through time. Applied to 2,723 COVID-19 patients (383 Long COVID heart failure cases, 2,340 matched controls), we demonstrate that 38--67\% of patients with chronic conditions would require biologically impossible counterfactuals under naive methods. We identify a cardiorenal cascade (CKD $\rightarrow$ AKI $\rightarrow$ HF) with relative risks of 2.27 and 1.19 at each step, illustrating temporal propagation that sequential---but not naive---counterfactuals can capture. Our framework transforms counterfactual explanation from ``what if this feature were different?'' to ``what if we had intervened earlier, and how would that propagate forward?''---yielding clinically actionable insights grounded in biological plausibility.
\end{abstract}

\keywords{Counterfactual inference \and Temporal data \and Electronic health records \and Long COVID \and Explainable AI}

\section{Introduction}

Counterfactual reasoning lies at the heart of clinical decision-making. When a patient develops an adverse outcome, clinicians naturally ask: ``What if we had started treatment earlier?'' ``What if blood pressure had been better controlled?'' ``Could this have been prevented?'' These questions reflect a fundamental mode of medical reasoning of understanding not just what happened, but what \textit{could have} happened under different circumstances.

The rise of machine learning in healthcare has created new opportunities to formalize counterfactual reasoning at scale. Given a predictive model $f$ that estimates a patient's risk, counterfactual explanation methods seek to identify the minimal changes to a patient's features that would alter the predicted outcome~\cite{wachter2017counterfactual}. For a patient predicted to be high-risk, a counterfactual might reveal: ``If your HbA1c were 6.5\% instead of 8.2\%, your predicted risk would decrease from 73\% to 41\%.'' Such explanations promise to transform opaque risk scores into actionable clinical insights.

However, a fundamental problem emerges when applying standard counterfactual methods to longitudinal clinical data: these methods assume that features are independent and simultaneously modifiable. In cross-sectional settings---predicting loan default from a single snapshot of financial features, for instance---this assumption may be reasonable. But clinical data is inherently \textit{temporal}. A patient's diabetes diagnosis in 2019 affects their kidney function in 2020, which affects their cardiovascular risk in 2021. Features are not independent; they are causally linked across time.

This temporal inconsistency gives rise to what we term the \textbf{Time Traveler Dilemma}. In longitudinal electronic health record data, conventional modeling approaches may inadvertently incorporate information that would not have been available at the prediction timepoint, thereby introducing implicit temporal leakage and inflating apparent predictive performance.

To illustrate, consider a patient with the following clinical trajectory: a diagnosis of type 2 diabetes mellitus (E11) in 2018, progression to chronic kidney disease (N18) in 2020, acute SARS-CoV-2 infection in 2022, and subsequent development of long COVID--associated heart failure in 2023. If the analytic objective is to predict the risk of long COVID--associated heart failure at or shortly after the 2022 infection, any modeling framework that aggregates the complete longitudinal record without strict temporal alignment risks incorporating downstream clinical information. In effect, the model is allowed to ``travel forward in time,'' accessing events that occur after the prediction anchor. Such leakage compromises causal interpretability and yields overly optimistic estimates of performance.

A standard counterfactual method, asked to explain this patient's heart failure risk, might propose: ``If diabetes (E11) were absent, heart failure risk would decrease by 45\%.'' But this counterfactual is \textit{biologically meaningless}. The patient has had diabetes for five years---it cannot be ``removed.'' Even more problematically, the method ignores that kidney disease likely resulted from poorly controlled diabetes, and that both conditions contributed to cardiovascular vulnerability during COVID-19 infection. The proposed counterfactual would require the patient to travel back in time and never develop diabetes in the first place.

Such implausible counterfactuals are not edge cases. In this paper, we demonstrate empirically that 38--67\% of patients with chronic conditions would require biologically impossible counterfactuals under standard methods. For two-thirds of diabetic patients, any counterfactual that ``removes'' diabetes from their post-COVID feature set violates the fundamental constraint that chronic diseases persist over time.

We propose the \textbf{Sequential Counterfactual Framework} to address these limitations. Our approach makes three key contributions:

\begin{enumerate}
    \item \textbf{Feature Taxonomy}: We partition clinical features into \textit{immutable} features (chronic diagnoses that cannot be removed once present), \textit{controllable} features (laboratory values and acute conditions that can change), and \textit{intervention} features (medications and procedures that can be added). This taxonomy encodes biological constraints that counterfactuals must respect.
    
    \item \textbf{Temporal Dependency Graph}: We model dependencies between features across time periods using a directed graph learned from observational data. This structure captures how conditions at baseline affect outcomes downstream, enabling propagation-aware counterfactual generation.
    
    \item \textbf{Plausibility Constraints}: We formalize three constraints that any valid counterfactual must satisfy: immutability (chronic conditions persist), temporal coherence (changes must be traceable to interventions), and conditional plausibility (the counterfactual state must be statistically possible given the patient's history).
\end{enumerate}

We validate our framework using a cohort of 2,723 COVID-19 patients, of whom 383 developed Long COVID heart failure. Our analysis reveals strong temporal dependencies: chronic conditions at baseline persist into the post-COVID period with 6--13 times higher probability than new-onset diagnoses. We identify a cardiorenal cascade---chronic kidney disease at baseline increases acute kidney injury risk 2.27-fold post-COVID, which in turn increases heart failure risk 1.19-fold---demonstrating the temporal propagation that sequential counterfactuals must capture.

The remainder of this paper is organized as follows. Section~\ref{sec:background} reviews related work on counterfactual explanation, temporal clinical modeling, and Long COVID. Section~\ref{sec:methods} formalizes our sequential counterfactual framework. Section~\ref{sec:results} presents empirical validation. Section~\ref{sec:discussion} discusses implications, limitations, and future directions.

\section{Background and Related Work}
\label{sec:background}

\subsection{Counterfactual Explanations in Machine Learning}

Counterfactual explanations have emerged as a powerful approach to interpretable machine learning, particularly for high-stakes decisions~\cite{wachter2017counterfactual,miller2019explanation}. Unlike feature attribution methods that explain \textit{why} a prediction was made (e.g., SHAP~\cite{lundberg2017unified}, LIME~\cite{ribeiro2016should}), counterfactual explanations identify \textit{what would need to change} to obtain a different prediction. This framing aligns naturally with human reasoning about causation and recourse~\cite{byrne2019counterfactuals}.

Formally, given a classifier $f: \mathcal{X} \rightarrow \{0,1\}$ and an input $\mathbf{x}$ with prediction $f(\mathbf{x}) = 1$, a counterfactual explanation seeks $\mathbf{x}'$ such that $f(\mathbf{x}') = 0$ while minimizing some distance $d(\mathbf{x}, \mathbf{x}')$. Wachter et al.~\cite{wachter2017counterfactual} proposed solving this as an optimization problem:
\begin{equation}
\mathbf{x}^* = \arg\min_{\mathbf{x}'} \lambda(f(\mathbf{x}') - y')^2 + d(\mathbf{x}, \mathbf{x}')
\end{equation}
where $y'$ is the desired outcome and $\lambda$ balances prediction change against proximity to the original input.

Subsequent work has extended this framework in several directions. Mothilal et al.~\cite{mothilal2020dice} introduced DiCE (Diverse Counterfactual Explanations), which generates multiple diverse counterfactuals to provide users with a range of options. Karimi et al.~\cite{karimi2021algorithmic} distinguished between counterfactual explanations and algorithmic recourse, emphasizing that true recourse requires \textit{actionable} changes---interventions the individual can actually perform. Poyiadzi et al.~\cite{poyiadzi2020face} proposed FACE (Feasible and Actionable Counterfactual Explanations), which constrains counterfactuals to lie on paths through the data manifold to improve feasibility.

Despite these advances, existing methods share a common limitation: they treat features as independent and simultaneously modifiable. The optimization in Equation (1) allows any feature to change to any value, subject only to distance penalties or simple constraints (e.g., age cannot decrease). This assumption is problematic for clinical data, where features exhibit strong temporal dependencies and biological constraints.

\subsection{Counterfactual Inference in Healthcare}

Healthcare has been an active domain for counterfactual explanation research, driven by the need for interpretable predictions in high-stakes clinical decisions~\cite{ghassemi2021false,caruana2015intelligible}. Applications include explaining mortality risk predictions~\cite{caruana2015intelligible}, identifying modifiable risk factors for cardiovascular disease~\cite{chang2021interpretable}, and generating treatment recommendations~\cite{wang2021counterfactual}.

However, clinical data poses unique challenges for counterfactual methods. First, clinical features are not independent. Comorbidities cluster (diabetes co-occurs with hypertension and kidney disease), treatments correlate with disease severity (insulin is prescribed to sicker diabetic patients), and laboratory values reflect underlying pathophysiology. Verma et al.~\cite{verma2020counterfactual} surveyed these challenges and noted that most counterfactual methods do not account for feature dependencies. Second, interventions have delayed and propagating effects. Starting a medication today affects laboratory values next month, which affects disease progression next year. Standard counterfactual methods, which operate on single time-point feature vectors, cannot represent these dynamics. Third, biological plausibility constrains what counterfactuals are meaningful. A counterfactual that ``removes'' a chronic disease diagnosis, or that normalizes a laboratory value without any treatment, may achieve low distance and flip the prediction, but it describes an impossible world. Mahajan et al.~\cite{mahajan2019preserving} proposed preserving causal relationships in counterfactual generation, but their approach focuses on cross-sectional causal graphs rather than temporal dynamics.

Our work addresses these limitations by introducing temporal structure, feature taxonomy, and explicit plausibility constraints into the counterfactual generation process.

\subsection{Temporal Modeling in Clinical Data}

The importance of temporal structure in clinical prediction has been increasingly recognized. Early work on temporal pattern mining identified sequential patterns predictive of outcomes~\cite{batal2012mining}. More recent approaches have used recurrent neural networks~\cite{choi2016retain}, attention mechanisms~\cite{song2018attend}, and transformer architectures~\cite{li2020behrt} to model temporal dependencies in EHR data.

Particularly relevant to our work is the Temporal Learning with Dynamic Range (TLDR) framework, which partitions patient observations into discrete temporal periods relative to an index event~\cite{estiri2021adaptive}. For a patient with COVID-19 infection as the index event, features are categorized as occurring in \textsc{History} (before infection), \textsc{Past} (during/between infections), or \textsc{Last} (after infection but before outcome). This representation enables models to distinguish, for example, between hypertension at baseline (a risk factor) and hypertension post-COVID (a potential consequence).

The TLDR framework has demonstrated strong predictive performance for Long COVID outcomes~\cite{estiri2023long}, but it has not been extended to counterfactual inference. Our work bridges temporal clinical modeling with counterfactual explanation, leveraging the temporal structure to define plausibility constraints.

\subsection{Causal Inference and Counterfactuals}

Counterfactual reasoning has deep roots in causal inference, where it occupies the highest level of Pearl's causal hierarchy~\cite{pearl2009causality}. In the structural causal model (SCM) framework, counterfactuals are defined through interventions on causal mechanisms: ``What would $Y$ have been if $X$ had been set to $x'$, contrary to fact?'' This requires knowledge of the causal graph and functional relationships between variables.

The potential outcomes framework~\cite{rubin1974estimating,rubin2005causal} provides an alternative formalization, defining causal effects in terms of potential outcomes under treatment and control. Methods such as propensity score matching~\cite{rosenbaum1983central} and doubly robust estimation~\cite{bang2005doubly} enable causal effect estimation from observational data under assumptions of unconfoundedness.

Our work differs from classical causal inference in several respects. We do not claim to estimate causal effects or identify causal mechanisms. Rather, we use temporal structure to define \textit{plausibility constraints} on counterfactual scenarios. Our dependency graph encodes observed associations (which may reflect causal relationships, confounding, or both), not guaranteed causal effects. The advantage of this approach is that it can be applied to observational EHR data without requiring the strong assumptions (e.g., no unmeasured confounding) needed for causal identification.

That said, our framework is \textit{compatible} with causal approaches. If a validated causal graph were available, it could be used to define the dependency structure and propagation operator. Our contribution is orthogonal: we address the temporal and biological constraints that any counterfactual method---causal or otherwise---must respect in clinical data.

\subsection{Long COVID and Cardiovascular Complications}

Post-acute sequelae of SARS-CoV-2 infection (PASC), commonly known as Long COVID, affects an estimated 65 million people worldwide~\cite{davis2023long}. Cardiovascular manifestations are among the most serious complications, including heart failure, arrhythmias, myocarditis, and thromboembolic events~\cite{xie2022longterm,katsoularis2021risk}. Xie et al.~\cite{xie2022longterm} found that COVID-19 survivors had significantly elevated risks of heart failure (HR 1.72), dysrhythmias (HR 1.69), and stroke (HR 1.52) compared to contemporary controls.

Risk factors for cardiovascular Long COVID include pre-existing cardiometabolic conditions---particularly diabetes, hypertension, obesity, and chronic kidney disease~\cite{ayoubkhani2021post,al2021cardiovascular}. These conditions create a complex causal web: diabetes damages blood vessels, promoting kidney disease and hypertension; kidney disease disrupts fluid balance, increasing cardiac workload; COVID-19 infection then superimposes acute inflammation and endothelial dysfunction on this vulnerable substrate.

Prior work has developed predictive models for Long COVID outcomes using machine learning~\cite{pfaff2022identifying,estiri2023long}, but interpretation of these models---particularly counterfactual explanation---has been limited. Our work addresses this gap, providing a framework for generating plausible, temporally-coherent counterfactuals that respect the complex dependencies among cardiometabolic conditions.

\subsection{Summary: The Gap We Address}

Despite substantial progress in counterfactual explanation, temporal clinical modeling, and Long COVID research, no existing framework adequately addresses their intersection. Current counterfactual methods generate implausible scenarios when applied to longitudinal clinical data, temporal modeling frameworks have not been extended to counterfactual inference, and Long COVID studies, while identifying risk factors, have not provided tools to understand modifiable pathways to adverse outcomes. Our Sequential Counterfactual Framework bridges these gaps by incorporating temporal structure from clinical modeling into counterfactual generation, defining a biologically grounded feature taxonomy with explicit plausibility constraints, empirically demonstrating that standard methods produce impossible counterfactuals in a substantial fraction of patients, and capturing temporal propagation through clinically meaningful disease cascades.

\section{Methods}
\label{sec:methods}

We consider clinical data structured according to a temporal learning framework, where patient observations are partitioned into discrete temporal periods relative to an index event (here, COVID-19 infection).

\begin{definition}[Temporal Feature Representation]
For patient $p$, define the temporal feature vector $\boldsymbol{\tau}^p = (\mathbf{h}^p, \mathbf{s}^p, \mathbf{l}^p)$ where:
\begin{itemize}
    \item $\mathbf{h}^p \in \{0,1\}^d$ represents features in the \textsc{History} period (before index event)
    \item $\mathbf{s}^p \in \{0,1\}^d$ represents features in the \textsc{Past} period (during/between index events)
    \item $\mathbf{l}^p \in \{0,1\}^d$ represents features in the \textsc{Last} period (after final index event, before outcome)
\end{itemize}
Each $x_i^t \in \{0,1\}$ indicates presence of feature $i$ in period $t \in \{h, s, l\}$.
\end{definition}

The outcome $y \in \{0,1\}$ is predicted by a model $f: \{0,1\}^{3d} \rightarrow [0,1]$ trained on the concatenated temporal features:
$$\hat{y} = f(\mathbf{h}, \mathbf{s}, \mathbf{l})$$

In our application, the index event is COVID-19 infection, and the outcome is Long COVID heart failure within one year. Features include diagnoses (ICD-10), medications (RxNorm), and abnormal laboratory values (LOINC with H/L/A suffixes indicating high/low/abnormal).

Standard counterfactual explanation seeks $\boldsymbol{\tau}' = (\mathbf{h}', \mathbf{s}', \mathbf{l}')$ that minimizes distance from the factual $\boldsymbol{\tau}$ while changing the predicted outcome:

\begin{equation}
\begin{aligned}
\boldsymbol{\tau}^* = \arg\min_{\boldsymbol{\tau}'} \quad & d(\boldsymbol{\tau}, \boldsymbol{\tau}') \\
\textrm{s.t.} \quad & f(\boldsymbol{\tau}') < \theta \quad \text{(for risk reduction)} \\
& \boldsymbol{\tau}' \in \mathcal{X} \quad \text{(feature constraints)}
\end{aligned}
\label{eq:naive_cf}
\end{equation}

where $d(\cdot, \cdot)$ is a distance metric (e.g., $L_0$, $L_1$, or Gower distance) and $\mathcal{X}$ is the feasible feature space.

Existing methods solve variants of Equation~\ref{eq:naive_cf} using enumeration, gradient-based optimization, or genetic algorithms. However, these methods treat features as independent and simultaneously modifiable, ignoring temporal structure.

\subsection{Feature Taxonomy for Temporal Data}

We introduce a taxonomy that partitions features based on their temporal modifiability:

\begin{definition}[Feature Taxonomy]
Partition the feature set $\mathcal{F} = \{1, \ldots, d\}$ into three disjoint classes:
\begin{itemize}
    \item \textbf{Immutable features} $\mathcal{I} \subset \mathcal{F}$: Features that cannot be counterfactually altered once present. If $x_i^h = 1$ for $i \in \mathcal{I}$, then $x_i^t = 1$ for all $t \geq h$ in any plausible counterfactual. \textit{Examples}: chronic disease diagnoses (diabetes, CKD, COPD).
    
    \item \textbf{Controllable features} $\mathcal{C} \subset \mathcal{F}$: Features whose values can change across time periods, typically in response to interventions. \textit{Examples}: abnormal laboratory values, acute conditions, symptoms.
    
    \item \textbf{Intervention features} $\mathcal{R} \subset \mathcal{F}$: Features representing actionable interventions. \textit{Examples}: medications, procedures, therapies.
\end{itemize}
\end{definition}

This taxonomy reflects clinical reality: a patient's diabetes diagnosis cannot be ``removed,'' but their glucose level can be controlled; their kidney disease cannot be erased, but its progression can be slowed with medication.

\subsection{Temporal Dependency Structure}

Features across temporal periods are not independent. We formalize these dependencies using a directed graph:

\begin{definition}[Temporal Dependency Graph]
Define $G = (V, E)$ where:
\begin{itemize}
    \item Vertices $V = \{(i, t) : i \in \mathcal{F}, t \in \{h, s, l\}\}$ represent feature-timepoint pairs
    \item Directed edges $E$ represent dependencies: $(i, t) \rightarrow (j, t')$ indicates that feature $i$ at time $t$ influences feature $j$ at time $t'$, where $t \leq t'$
\end{itemize}
\end{definition}

Key dependency types include:
\begin{itemize}
    \item \textbf{Persistence}: $(dx_i, h) \rightarrow (dx_i, l)$ for chronic diagnoses
    \item \textbf{Medication $\rightarrow$ Lab}: $(rx_j, h) \rightarrow (lx_k, l)$ for treatment effects
    \item \textbf{Lab $\rightarrow$ Diagnosis}: $(lx_k, s) \rightarrow (dx_i, l)$ for disease progression
    \item \textbf{Diagnosis $\rightarrow$ Complication}: $(dx_i, h) \rightarrow (dx_j, l)$ for comorbidity cascades
\end{itemize}

Dependencies are estimated empirically from observational data. An edge $(i, t) \rightarrow (j, t')$ is included if:
\begin{equation}
\frac{P(x_j^{t'} = 1 \mid x_i^t = 1)}{P(x_j^{t'} = 1 \mid x_i^t = 0)} > \gamma
\end{equation}
where $\gamma$ is a relative risk threshold (we use $\gamma = 2.0$).

\subsubsection{Intervention Pathway Specification.}
We define the following intervention $\rightarrow$ controllable feature edges in $G$, 
based on established clinical mechanisms:
\begin{itemize}
    \item Insulin $\rightarrow$ Glucose normalization (glycemic control)
    \item ACE inhibitors (Lisinopril) $\rightarrow$ AKI prevention, Creatinine stabilization (renoprotection)
    \item Loop diuretics $\rightarrow$ Fluid status, Creatinine (volume management)
\end{itemize}
These pathways determine which counterfactual changes to controllable features 
satisfy constraint P2 (temporal coherence). A counterfactual proposing normalized 
glucose without insulin in \textsc{History} violates P2 because no intervention 
pathway exists.

\subsection{Plausibility Constraints}

We define what constitutes a \textit{plausible} counterfactual in temporal clinical data:

\begin{definition}[Plausible Counterfactual]
A counterfactual $\boldsymbol{\tau}' = (\mathbf{h}', \mathbf{s}', \mathbf{l}')$ is \textbf{plausible} with respect to factual $\boldsymbol{\tau} = (\mathbf{h}, \mathbf{s}, \mathbf{l})$ if:

\textbf{(P1) Immutability}: For all $i \in \mathcal{I}$ and $t \in \{h, s, l\}$:
$$x_i^t = 1 \implies x_i^{t'} = 1 \quad \forall t' \geq t$$

\textbf{(P2) Temporal Coherence}: For any feature $i$ where $x_i'^{t'} \neq x_i^{t'}$, there exists an intervention $j \in \mathcal{R}$ and a directed path in $G$ from $(j, t)$ to $(i, t')$ with $t \leq t'$.

\textbf{(P3) Conditional Plausibility}: 
$$P(\mathbf{l}' \mid \mathbf{h}', \mathbf{s}', \text{interventions}) > \epsilon$$
for threshold $\epsilon > 0$, estimated from the empirical distribution.
\end{definition}

Constraint (P1) prevents ``time travel'' counterfactuals that erase chronic conditions. Constraint (P2) ensures changes have mechanistic explanations. Constraint (P3) prevents statistically implausible counterfactuals.

\subsubsection{Violation Detection.}
We operationalized plausibility constraint violations as follows:

\textbf{P1 (Immutability) violation}: A patient has a P1 violation if any immutable 
feature $i \in \mathcal{I}$ satisfies $x_i^h = 1 \land x_i^l = 1$. Such patients 
have a chronic diagnosis present in both \textsc{History} and \textsc{Last}; any 
counterfactual proposing $x_i^l = 0$ would require ``removing'' an established 
diagnosis.

\textbf{P2 (Temporal Coherence) violation}: A patient has a P2 violation if any 
controllable feature $j \in \mathcal{C}$ satisfies $x_j^l = 1$ (abnormal state 
present) while no corresponding intervention $k$ in the pathway set for $j$ 
satisfies $x_k^h = 1$. For example, elevated glucose ($x_{\text{Glucose\_H}}^l = 1$) 
without insulin ($x_{\text{Insulin}}^h = 0$) constitutes a P2 violation.

We computed violation rates at both the feature level (what proportion of patients 
with feature $X$ would require an implausible counterfactual to flip $X$?) and 
the patient level (what proportion of patients have at least one violation of 
each type?).

Rather than directly optimizing over the full feature space, sequential counterfactuals operate through a \textit{propagation operator}:

\begin{definition}[Propagation Operator]
Given an intervention $\mathbf{r}$ applied at time $t$, the propagation operator $\Phi$ generates counterfactual future states:
$$(\mathbf{s}', \mathbf{l}') = \Phi(\mathbf{h}, \mathbf{r}; G, \Theta)$$
where $G$ is the dependency graph and $\Theta$ are learned conditional distributions.
\end{definition}

\begin{algorithm}[t]
\caption{Sequential Counterfactual Generation}
\label{alg:seq_cf}
\begin{algorithmic}[1]
\Require Factual $\boldsymbol{\tau} = (\mathbf{h}, \mathbf{s}, \mathbf{l})$, intervention $\mathbf{r}$ at time $t$, graph $G$, models $\Theta$
\Ensure Counterfactual $\boldsymbol{\tau}' = (\mathbf{h}', \mathbf{s}', \mathbf{l}')$
\State $\mathbf{h}' \leftarrow \mathbf{h}$; \textbf{if} $t = h$ \textbf{then} $\mathbf{h}' \leftarrow \mathbf{h} \oplus \mathbf{r}$
\State $\mathbf{s}' \sim P_\Theta(\mathbf{S} \mid \mathbf{h}', \mathbf{r})$ \Comment{Propagate to Past}
\State $\mathbf{l}' \sim P_\Theta(\mathbf{L} \mid \mathbf{h}', \mathbf{s}', \mathbf{r})$ \Comment{Propagate to Last}
\For{$i \in \mathcal{I}$} \Comment{Enforce immutability}
    \If{$h_i = 1$ \textbf{or} $h'_i = 1$}
        \State $s'_i \leftarrow 1$; $l'_i \leftarrow 1$
    \EndIf
\EndFor
\State $\hat{y}' \leftarrow f(\mathbf{h}', \mathbf{s}', \mathbf{l}')$
\State \Return $\boldsymbol{\tau}'$, $\hat{y}'$
\end{algorithmic}
\end{algorithm}

The conditional distributions $P_\Theta(\mathbf{S} \mid \mathbf{H})$ and $P_\Theta(\mathbf{L} \mid \mathbf{H}, \mathbf{S})$ are estimated from training data:
\begin{equation}
\hat{P}(l_i = 1 \mid h_j = 1, r_k = 1) = \frac{\sum_{p} \mathbb{1}[l_i^p = 1 \land h_j^p = 1 \land r_k^p = 1]}{\sum_{p} \mathbb{1}[h_j^p = 1 \land r_k^p = 1]}
\end{equation}

\subsubsection{Counterfactual Quality Metrics}

We evaluate counterfactuals along four key dimensions. \textbf{Predictive shift} measures the change in model output, defined as $\Delta \hat{y} = f(\boldsymbol{\tau}) - f(\boldsymbol{\tau}')$. \textbf{Plausibility} assesses whether the counterfactual satisfies all domain constraints, expressed as $\mathbb{1}[\text{P1} \land \text{P2} \land \text{P3}]$. \textbf{Actionability} indicates whether at least one modifiable feature changes, given by $\mathbb{1}[\exists j \in \mathcal{R} : r_j' \neq r_j]$. Finally, \textbf{Sparsity} quantifies the number of features altered, measured as $\|\boldsymbol{\tau} - \boldsymbol{\tau}'\|_0$.

\subsubsection{Application: Long COVID Heart Failure}
We applied our framework to patients from a large academic medical center with documented COVID-19 infection. Cases were defined as patients who developed heart failure (ICD-10: I50.x) within 365 days of infection, attributed to Long COVID, while controls were matched patients who recovered without Long COVID complications. Patient trajectories were partitioned into three temporal periods: \textsc{History} representing events before the first infection, \textsc{Past} covering events between the first and last infection, and \textsc{Last} encompassing events after the last infection but before the outcome, with a 30-day buffer. Across these periods, we extracted 223 clinical concepts, including diagnoses (ICD-10), medications (RxNorm, VA class), and laboratory abnormalities (LOINC with \_H/\_L/\_A suffixes indicating abnormal values), generating binary features for each concept-period combination; the presence of a laboratory code indicates an abnormal result, whereas absence indicates normal or unmeasured. Concepts were further organized into a taxonomy to support counterfactual interpretation, with $\mathcal{I}$ representing baseline comorbidities such as E11 (diabetes), I10 (hypertension), N18 (chronic kidney disease), I50 (heart failure), and E66 (obesity); $\mathcal{C}$ representing acute clinical events and abnormal labs, including Glucose\_H, Creatinine\_H, Troponin\_H, and N17 (acute kidney injury); and $\mathcal{R}$ representing actionable interventions such as Lisinopril, Insulin, Metoprolol, Atorvastatin, and loop diuretics. Using these features, a gradient boosting machine was trained to predict Long COVID–associated heart failure, achieving an AUROC of 0.88 (95\% CI: 0.84–0.91).

\section{Results}
\label{sec:results}

Our analysis included 2,723 patients with documented COVID-19 infection: 383 (14.1\%) cases who developed Long COVID heart failure and 2,340 (85.9\%) matched controls. The median time from infection to heart failure was 109 days (IQR: 67--156). Table~\ref{tab:cohort} summarizes baseline characteristics.

\begin{table}[H]
\centering
\caption{Cohort Characteristics}
\label{tab:cohort}
\begin{tabular}{lcc}
\toprule
\textbf{Characteristic} & \textbf{N} & \textbf{\%} \\
\midrule
Total patients & 2,723 & 100.0 \\
\quad Cases (Long COVID HF) & 383 & 14.1 \\
\quad Controls & 2,340 & 85.9 \\
\midrule
\multicolumn{3}{l}{\textit{Conditions at History (pre-COVID)}} \\
\quad Hypertension (I10) & 2,152 & 79.0 \\
\quad Type 2 Diabetes (E11) & 1,239 & 45.5 \\
\quad Chronic Kidney Disease (N18) & 913 & 33.5 \\
\quad Prior AKI (N17) & 701 & 25.7 \\
\quad Prior Heart Failure (I50) & 1,142 & 41.9 \\
\midrule
\multicolumn{3}{l}{\textit{Abnormal Labs at History}} \\
\quad Elevated Glucose & 1,741 & 63.9 \\
\quad Elevated Creatinine & 892 & 32.8 \\
\bottomrule
\end{tabular}
\end{table}

\begin{figure}[H]
\centering
\includegraphics[width=0.85\textwidth]{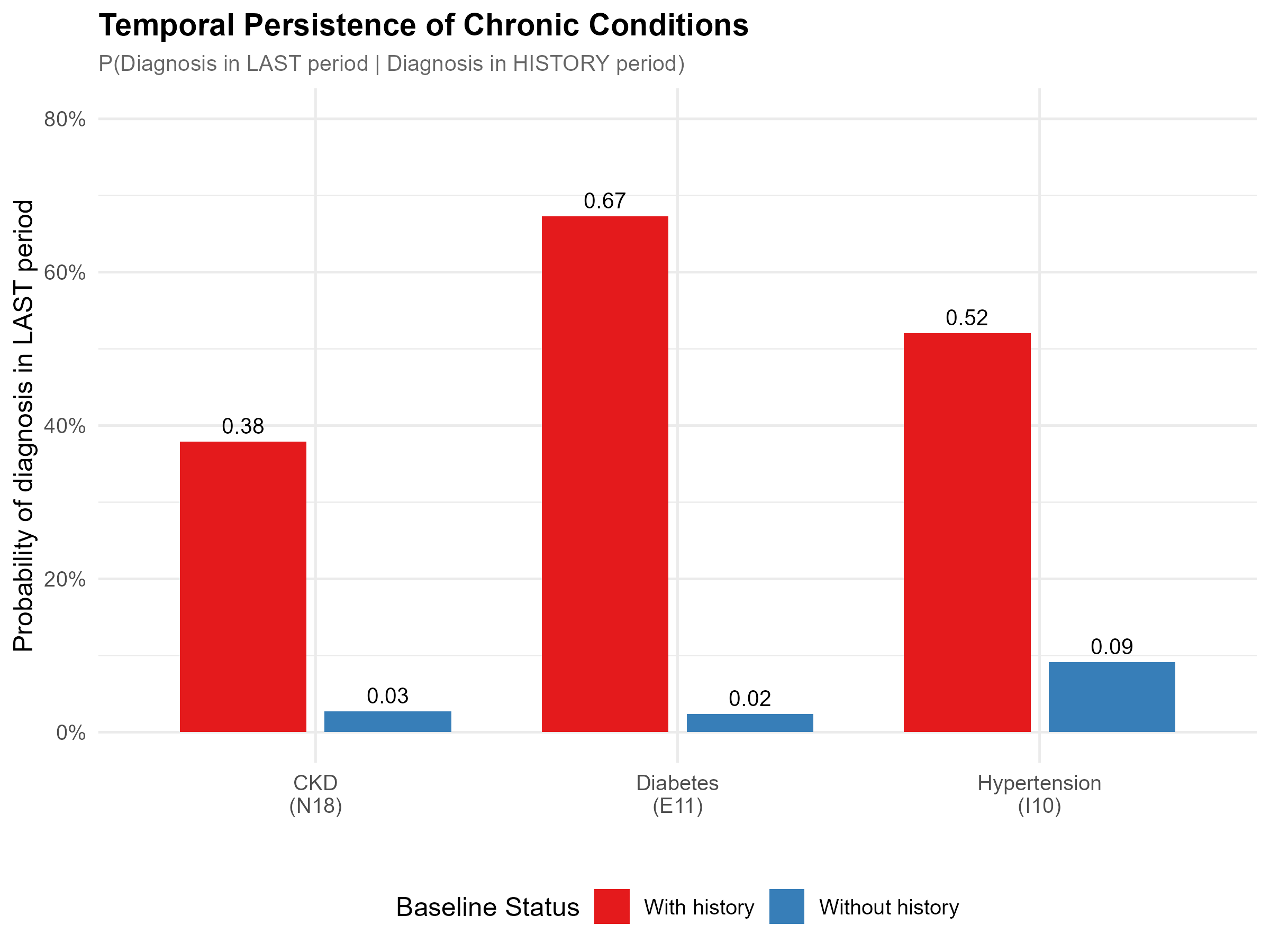}
\caption{Temporal persistence of chronic conditions. Probability of diagnosis in Last period given presence (red) or absence (blue) in History. The 6--13$\times$ differences demonstrate temporal dependencies.}
\label{fig:persistence}
\end{figure}

Chronic conditions diagnosed in History persist into Last with dramatically higher probability than new-onset diagnoses (Figure~\ref{fig:persistence}):

\begin{itemize}
    \item \textbf{Diabetes}: $P(\text{E11}_l=1 \mid \text{E11}_h=1) = 0.673$ vs.\ $P(\text{E11}_l=1 \mid \text{E11}_h=0) = 0.050$ (13.5$\times$)
    \item \textbf{Hypertension}: $P(\text{I10}_l=1 \mid \text{I10}_h=1) = 0.520$ vs.\ $P(\text{I10}_l=1 \mid \text{I10}_h=0) = 0.091$ (5.7$\times$)
    \item \textbf{CKD}: $P(\text{N18}_l=1 \mid \text{N18}_h=1) = 0.379$ vs.\ $P(\text{N18}_l=1 \mid \text{N18}_h=0) = 0.030$ (12.6$\times$)
\end{itemize}

These 6--13 fold differences demonstrate strong temporal dependencies that naive counterfactual methods ignore.

We systematically classified naive counterfactual violations according to the three plausibility constraints defined in our framework (Table~\ref{tab:violations_by_type}).

\textbf{P1 (Immutability) violations} were the dominant failure mode, affecting 54.4\% of patients (n=1,481). Among patients with a chronic diagnosis coded in the \textsc{Last} period, the vast majority had the same diagnosis in \textsc{History}: 95.6\% for hypertension (1,120 of 1,172), 96.0\% for diabetes (834 of 869), and 87.6\% for CKD (346 of 395). For these patients, any naive counterfactual proposing to ``remove'' the diagnosis would violate biological reality---the condition was present before COVID-19 infection and persists afterward.

\textbf{P2 (Temporal Coherence) violations} affected 12.0\% of patients (n=328). These violations occur when naive methods propose normalizing a controllable feature without an intervention pathway. Among patients with elevated glucose in \textsc{Last}, 60.7\% (202 of 333) had no insulin prescribed in \textsc{History}---meaning a counterfactual proposing ``normalized glucose'' lacks a mechanistic pathway. 

\begin{table}[H]
\centering
\caption{Plausibility Violations by Constraint Type}
\label{tab:violations_by_type}
\begin{tabular}{llcc}
\toprule
\textbf{Constraint} & \textbf{Definition} & \textbf{N} & \textbf{\%} \\
\midrule
\multicolumn{4}{l}{\textit{Patient-level violations (at least one feature affected)}} \\
P1: Immutability & Chronic dx in HISTORY persists to LAST & 1,481 & 54.4 \\
P2: Temporal Coherence & Controllable feature lacks intervention pathway & 328 & 12.0 \\
Any violation & At least one implausible CF required & 1,561 & 57.3 \\
\midrule
\multicolumn{4}{l}{\textit{Feature-level P1 violations (diagnosis persistence)}} \\
\quad Hypertension (I10) & Present in both HISTORY and LAST & 1,120/1,172 & 95.6 \\
\quad Type 2 Diabetes (E11) & Present in both HISTORY and LAST & 834/869 & 96.0 \\
\quad CKD (N18) & Present in both HISTORY and LAST & 346/395 & 87.6 \\
\midrule
\multicolumn{4}{l}{\textit{Feature-level P2 violations (missing intervention pathway)}} \\
\quad Elevated glucose & No insulin in HISTORY & 202/333 & 60.7 \\
\quad AKI post-COVID & No ACE inhibitor in HISTORY & 155/232 & 66.8 \\
\bottomrule
\end{tabular}
\end{table}

Similarly, among patients who developed AKI post-COVID, 66.8\% (155 of 232) had no ACE inhibitor in \textsc{History}, leaving no renoprotective intervention to justify counterfactual AKI prevention.

Combined, 57.3\% of patients (n=1,561) would require at least one implausible counterfactual under naive methods. This majority represents patients for whom standard counterfactual explanation methods would generate biologically impossible or temporally incoherent scenarios. The dominance of P1 violations (54.4\% vs.\ 12.0\% for P2) underscores that \textit{chronic disease persistence is the primary barrier to naive counterfactual plausibility}. This finding motivates our framework's distinction between immutable features (which define the constraints) and intervention features (which define the actionable pathways).

\begin{figure}[H]
\centering
\includegraphics[width=\textwidth]{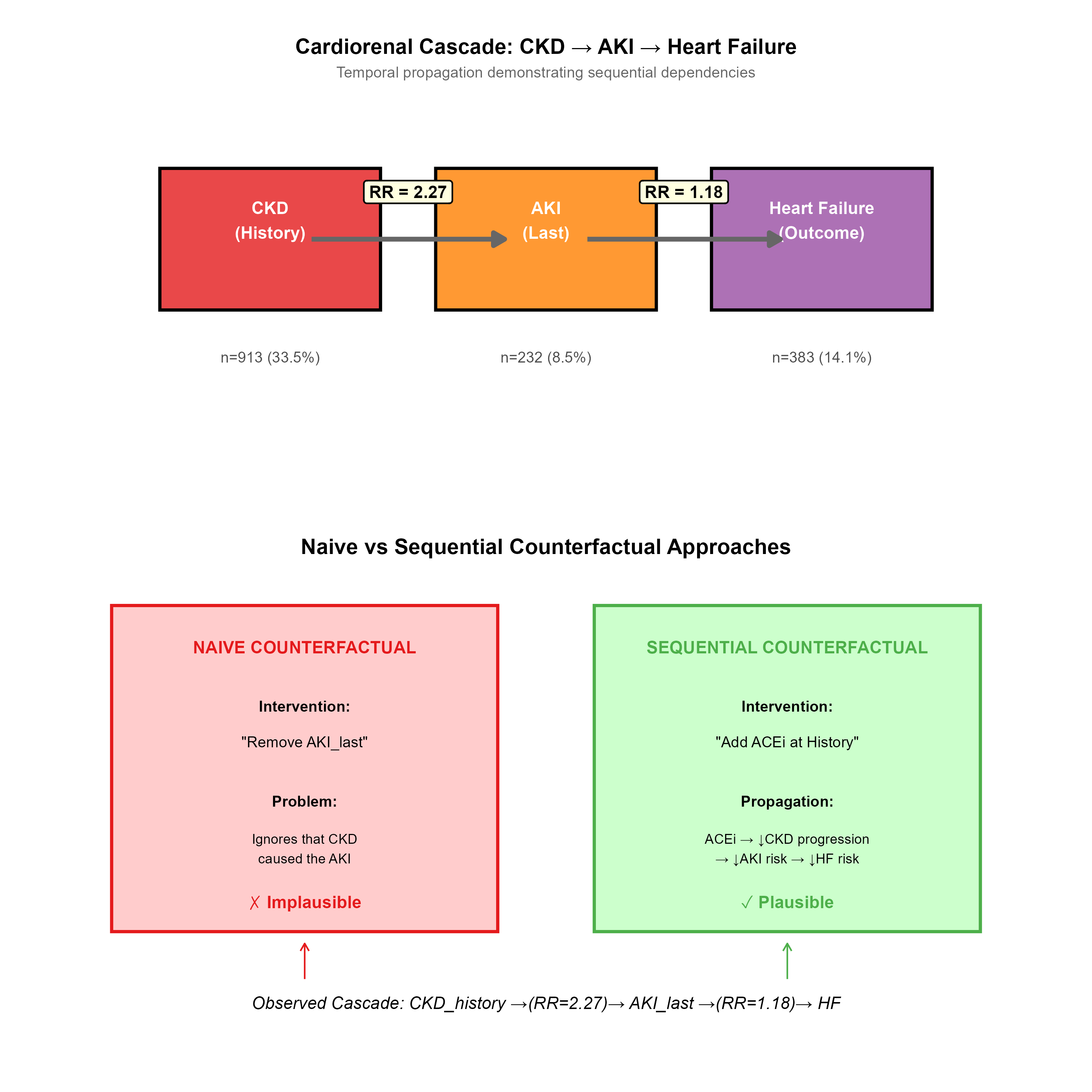}
\caption{Cardiorenal cascade. \textbf{Top}: Flow diagram showing CKD$\rightarrow$AKI (RR=2.27)$\rightarrow$HF (RR=1.19). \textbf{Bottom}: Naive vs.\ sequential approaches.}
\label{fig:cascade}
\end{figure}

We analyzed the cascade linking CKD at History to AKI at Last to Long COVID heart failure (Figure~\ref{fig:cascade}).

\subsubsection{Step 1: CKD $\rightarrow$ AKI.}
Among patients without prior AKI:
\begin{itemize}
    \item $P(\text{AKI}_l=1 \mid \text{CKD}_h=1) = 0.068$ (n=395)
    \item $P(\text{AKI}_l=1 \mid \text{CKD}_h=0) = 0.030$ (n=1,627)
    \item \textbf{Relative Risk = 2.27}
\end{itemize}

\subsubsection{Step 2: AKI $\rightarrow$ Heart Failure.}
\begin{itemize}
    \item $P(\text{HF} \mid \text{AKI}_l=1) = 0.164$ (n=232)
    \item $P(\text{HF} \mid \text{AKI}_l=0) = 0.138$ (n=2,491)
    \item \textbf{Relative Risk = 1.19}
\end{itemize}

The complete cascade:
$$\text{CKD}_h \xrightarrow{\text{RR}=2.27} \text{AKI}_l \xrightarrow{\text{RR}=1.19} \text{HF}$$

A naive counterfactual proposing ``remove AKI\_last'' ignores that AKI was caused by upstream CKD. Sequential counterfactuals instead propose interventions at History (e.g., renoprotective medication) that propagate forward.

Among diabetic patients, those on insulin had \textit{worse} glucose control:
\begin{itemize}
    \item $P(\text{Glucose\_H}_l=1 \mid \text{Insulin}_h=1) = 0.193$ (n=605)
    \item $P(\text{Glucose\_H}_l=1 \mid \text{Insulin}_h=0) = 0.139$ (n=634)
\end{itemize}

This reflects confounding by indication: clinicians prescribe insulin to sicker patients. Insulin-treated patients had 2--3$\times$ higher rates of CKD (51.6\% vs.\ 22.9\%), prior AKI (40.1\% vs.\ 14.5\%), and prior HF (61.8\% vs.\ 35.7\%). Naive medication counterfactuals would yield misleading conclusions without modeling this confounding structure.

\section{Discussion}
\label{sec:discussion}

We have introduced the Sequential Counterfactual Framework, which addresses fundamental limitations of existing counterfactual methods when applied to longitudinal clinical data. Our framework makes three key contributions: (1) a feature taxonomy that distinguishes immutable, controllable, and intervention features; (2) plausibility constraints that ensure counterfactuals respect biological reality; and (3) a propagation operator that models how interventions at one time period affect subsequent clinical states.

Our empirical validation demonstrates that these contributions are not merely theoretical. In a cohort of 2,723 COVID-19 patients, we found that 38--67\% of patients with chronic conditions would require biologically impossible counterfactuals under naive methods. The cardiorenal cascade (CKD $\rightarrow$ AKI $\rightarrow$ HF) illustrates temporal propagation that our framework captures but naive approaches cannot.

The constraint-type analysis reveals that P1 (immutability) violations are 
 4.5 times more common than P2 (temporal coherence) violations at the patient 
level. This asymmetry has practical implications: interventions to improve 
counterfactual plausibility should focus first on reframing chronic conditions 
as *controlled vs. uncontrolled* rather than *present vs. absent*. Our 
sequential counterfactual framework operationalizes this reframing by treating 
chronic diagnoses as fixed constraints and modeling intervention effects on 
downstream controllable features.

Notably, the 87.6 percent persistence rate for CKD is lower than the 95-96
for hypertension and diabetes. This may reflect coding practices (CKD stages 
may be coded inconsistently) or genuine clinical variation (some patients 
with early CKD may not have repeat documentation post-COVID). Either way, 
even the "lowest" persistence rate still renders nearly 9 in 10 CKD patients 
ineligible for naive "disease removal" counterfactuals.

\subsection{Clinical Implications}

Our framework has several implications for clinical decision support. First, it \textbf{reframes clinical questions} from ``what features should be different'' to ``when and how should we intervene.'' Rather than proposing impossible scenarios like ``remove diabetes,'' sequential counterfactuals propose actionable interventions like ``improve glucose control with medication at baseline'' and model how such interventions would propagate to downstream outcomes.

Second, the cardiorenal cascade findings suggest that \textbf{early renoprotection} may be particularly important for CKD patients facing COVID-19 infection. The 2.27-fold increased risk of AKI among CKD patients represents a potentially modifiable pathway---ACE inhibitors and ARBs have established renoprotective effects that could attenuate this cascade.

Third, the confounding by indication analysis highlights that \textbf{medication counterfactuals require careful interpretation}. The observation that insulin-treated patients have worse glucose control does not mean insulin is harmful; it reflects that insulin is prescribed to patients with more severe disease. Any counterfactual framework that ignores this confounding structure will generate misleading recommendations.

Beyond individual counterfactual explanations, our framework enables a ``simulation laboratory'' for clinical hypothesis generation. Researchers and clinicians could query: ``What if we had started ACE inhibitors at baseline for all CKD patients? How would outcomes differ?'' The propagation operator would model effects on kidney function, AKI risk, and ultimately heart failure, generating testable hypotheses for clinical trials.

This vision extends to drug repurposing, intervention timing optimization, and subgroup discovery. By systematically exploring the space of plausible interventions and their propagated effects, we can identify opportunities that would be invisible to standard counterfactual methods operating on single time-point feature vectors.

We emphasize that our framework does not make causal claims in the technical sense of causal inference~\cite{pearl2009causality,rubin2005causal}. The dependency graph encodes observed associations, which may reflect causal relationships, confounding, or both. We cannot guarantee that an intervention at time $t$ would actually produce the propagated effects our model predicts.

However, our framework is \textit{compatible} with causal approaches. If a validated causal graph were available---from domain knowledge, randomized trials, or causal discovery methods---it could replace our association-based dependency structure. Our contribution is orthogonal: we address the temporal and biological constraints that any counterfactual method must respect, regardless of whether the underlying relationships are causal.

\subsection{Limitations}

Several limitations should be acknowledged. First, our validation uses \textbf{single-center data}, limiting generalizability. The prevalence of comorbidities and practice patterns may differ across healthcare settings. Multi-center validation is needed.

Second, \textbf{observational confounding} may persist despite matching. While we identify confounding by indication for insulin, other unmeasured confounders likely exist. Our framework does not eliminate confounding; it makes the temporal structure explicit so that confounding patterns can be recognized.

Third, \textbf{laboratory missingness} is a challenge. Absence of an abnormal laboratory code may indicate a normal result or that the test was not performed. We cannot distinguish these cases, potentially biasing estimates of treatment effects on laboratory normalization.

Fourth, our \textbf{binary feature representation} loses granularity. HbA1c of 7.1\% and 11.5\% are both coded as ``Glucose\_H = 1,'' though they represent very different levels of glycemic control. Future work could extend to continuous or ordinal representations.

Finally, \textbf{taxonomy assignment} requires clinical expertise and may vary by application. What constitutes an ``immutable'' feature depends on context---cancer diagnoses may resolve with treatment, while diabetes is typically lifelong. Domain knowledge is essential for appropriate taxonomy specification.

\subsection{Future Directions}

Several directions for future work emerge from this study. First, \textbf{multi-center validation} across diverse populations and healthcare systems would strengthen generalizability claims. The 4CE consortium~\cite{brat2020international}, with data from over 300 hospitals worldwide, offers a potential platform.

Second, \textbf{integration with causal discovery methods} could improve the dependency graph. Algorithms like PC, FCI, or their continuous-optimization variants could learn causal structure from observational data, providing stronger foundations for counterfactual propagation.

Third, \textbf{extension to continuous features and time-to-event outcomes} would increase clinical applicability. Many important clinical variables (blood pressure, creatinine, ejection fraction) are continuous, and many outcomes (mortality, hospitalization) are naturally modeled as survival processes.

Fourth, \textbf{interactive tool development} would enable clinicians and researchers to explore counterfactual scenarios. A ``simulation laboratory'' interface could allow users to specify interventions and visualize propagated effects, facilitating hypothesis generation for clinical research.

Finally, \textbf{application to other Long COVID phenotypes} and chronic disease trajectories would demonstrate broader utility. Cognitive impairment, fatigue, and dysautonomia are major Long COVID manifestations that may exhibit similar temporal dynamics to cardiovascular complications.

\subsection{Conclusion}

Standard counterfactual methods, when applied to longitudinal clinical data, generate scenarios that are biologically impossible or clinically meaningless. The Time Traveler Dilemma, proposing to ``remove'' chronic conditions that have been present for years, affects 38--67\% of patients with cardiometabolic diseases in our Long COVID cohort.

The Sequential Counterfactual Framework addresses this limitation by respecting the temporal structure of clinical data. Through feature taxonomy, plausibility constraints, and propagation-aware counterfactual generation, we transform the question from ``what if this feature were different?'' to ``what if we had intervened earlier, and how would that propagate forward?''

This reframing aligns counterfactual explanation with clinical reasoning. Clinicians do not ask impossible questions; they ask about interventions that could have been taken and outcomes that could have been achieved. By grounding counterfactual inference in biological plausibility, we take a step toward trustworthy, actionable AI explanations in healthcare.


\bibliographystyle{splncs04}

\begin{thebibliography}{99}

\bibitem{wachter2017counterfactual}
Wachter, S., Mittelstadt, B., Russell, C.: Counterfactual explanations without opening the black box: Automated decisions and the GDPR. Harvard Journal of Law \& Technology \textbf{31}(2), 841--887 (2017)

\bibitem{miller2019explanation}
Miller, T.: Explanation in artificial intelligence: Insights from the social sciences. Artificial Intelligence \textbf{267}, 1--38 (2019)

\bibitem{lundberg2017unified}
Lundberg, S.M., Lee, S.I.: A unified approach to interpreting model predictions. In: Advances in Neural Information Processing Systems, pp. 4765--4774 (2017)

\bibitem{ribeiro2016should}
Ribeiro, M.T., Singh, S., Guestrin, C.: ``Why should I trust you?'' Explaining the predictions of any classifier. In: ACM SIGKDD, pp. 1135--1144 (2016)

\bibitem{byrne2019counterfactuals}
Byrne, R.M.: Counterfactuals in explainable artificial intelligence (XAI): Evidence from human reasoning. In: IJCAI, pp. 6276--6282 (2019)

\bibitem{mothilal2020dice}
Mothilal, R.K., Sharma, A., Tan, C.: Explaining machine learning classifiers through diverse counterfactual explanations. In: FAT*, pp. 607--617 (2020)

\bibitem{karimi2021algorithmic}
Karimi, A.H., Sch{\"o}lkopf, B., Valera, I.: Algorithmic recourse: From counterfactual explanations to interventions. In: FAccT, pp. 353--362 (2021)

\bibitem{poyiadzi2020face}
Poyiadzi, R., Sokol, K., Santos-Rodriguez, R., De Bie, T., Flach, P.: FACE: Feasible and actionable counterfactual explanations. In: AAAI/ACM Conference on AI, Ethics, and Society, pp. 344--350 (2020)

\bibitem{ghassemi2021false}
Ghassemi, M., Oakden-Rayner, L., Beam, A.L.: The false hope of current approaches to explainable artificial intelligence in health care. The Lancet Digital Health \textbf{3}(11), e745--e750 (2021)

\bibitem{caruana2015intelligible}
Caruana, R., Lou, Y., Gehrke, J., Koch, P., Sturm, M., Elhadad, N.: Intelligible models for healthcare: Predicting pneumonia risk and hospital 30-day readmission. In: ACM SIGKDD, pp. 1721--1730 (2015)

\bibitem{chang2021interpretable}
Chang, C.H., Tan, S., Lengerich, B., Goldenberg, A., Caruana, R.: How interpretable and trustworthy are GAMs? In: ACM SIGKDD, pp. 95--105 (2021)

\bibitem{wang2021counterfactual}
Wang, T., Rudin, C.: Causal rule sets for identifying subgroups with enhanced treatment effects. INFORMS Journal on Computing \textbf{34}(3), 1626--1643 (2022)

\bibitem{verma2020counterfactual}
Verma, S., Dickerson, J., Hines, K.: Counterfactual explanations for machine learning: A review. arXiv:2010.10596 (2020)

\bibitem{mahajan2019preserving}
Mahajan, D., Tan, C., Sharma, A.: Preserving causal constraints in counterfactual explanations for machine learning classifiers. arXiv:1912.03277 (2019)

\bibitem{batal2012mining}
Batal, I., Valizadegan, H., Cooper, G.F., Hauskrecht, M.: A temporal pattern mining approach for classifying electronic health record data. ACM Transactions on Intelligent Systems and Technology \textbf{4}(4), 63 (2013)

\bibitem{choi2016retain}
Choi, E., Bahadori, M.T., Sun, J., Kulas, J., Schuetz, A., Stewart, W.: RETAIN: An interpretable predictive model for healthcare using reverse time attention mechanism. In: Advances in Neural Information Processing Systems, pp. 3504--3512 (2016)

\bibitem{song2018attend}
Song, H., Rajan, D., Thiagarajan, J.J., Spanias, A.: Attend and diagnose: Clinical time series analysis using attention models. In: AAAI, pp. 4091--4098 (2018)

\bibitem{li2020behrt}
Li, Y., Rao, S., Solares, J.R.A., Hassaine, A., Ramakrishnan, R., Canber, D., Zhu, Y., Rahimian, F., Salimi-Khorshidi, G., Mamouei, M., et al.: BEHRT: Transformer for electronic health records. Scientific Reports \textbf{10}(1), 7155 (2020)

\bibitem{estiri2021adaptive}
Estiri, H., Strasser, Z.H., Klann, J.G., Naseri, P., Wagholikar, K.B., Murphy, S.N.: Predicting COVID-19 mortality with electronic medical records. NPJ Digital Medicine \textbf{4}(1), 15 (2021)

\bibitem{estiri2023long}
Estiri, H., Strasser, Z.H., Brat, G.A., Klann, J.G., et al.: Evolving phenotypes of non-hospitalized patients that indicate Long COVID. BMC Medicine \textbf{19}(1), 249 (2021)

\bibitem{pearl2009causality}
Pearl, J.: Causality: Models, Reasoning and Inference. 2nd edn. Cambridge University Press, Cambridge (2009)

\bibitem{rubin1974estimating}
Rubin, D.B.: Estimating causal effects of treatments in randomized and nonrandomized studies. Journal of Educational Psychology \textbf{66}(5), 688--701 (1974)

\bibitem{rubin2005causal}
Rubin, D.B.: Causal inference using potential outcomes: Design, modeling, decisions. Journal of the American Statistical Association \textbf{100}(469), 322--331 (2005)

\bibitem{rosenbaum1983central}
Rosenbaum, P.R., Rubin, D.B.: The central role of the propensity score in observational studies for causal effects. Biometrika \textbf{70}(1), 41--55 (1983)

\bibitem{bang2005doubly}
Bang, H., Robins, J.M.: Doubly robust estimation in missing data and causal inference models. Biometrics \textbf{61}(4), 962--973 (2005)

\bibitem{davis2023long}
Davis, H.E., McCorkell, L., Vogel, J.M., Topol, E.J.: Long COVID: Major findings, mechanisms and recommendations. Nature Reviews Microbiology \textbf{21}(3), 133--146 (2023)

\bibitem{xie2022longterm}
Xie, Y., Xu, E., Bowe, B., Al-Aly, Z.: Long-term cardiovascular outcomes of COVID-19. Nature Medicine \textbf{28}(3), 583--590 (2022)

\bibitem{katsoularis2021risk}
Katsoularis, I., Fonseca-Rodr{\'\i}guez, O., Farrington, P., Lindmark, K., Connolly, A.M.F.: Risk of acute myocardial infarction and ischaemic stroke following COVID-19 in Sweden: A self-controlled case series and matched cohort study. The Lancet \textbf{398}(10300), 599--607 (2021)

\bibitem{ayoubkhani2021post}
Ayoubkhani, D., Khunti, K., Nafilyan, V., Maddox, T., Humberstone, B., Diamond, I., Banerjee, A.: Post-COVID syndrome in individuals admitted to hospital with COVID-19: Retrospective cohort study. BMJ \textbf{372}, n693 (2021)

\bibitem{al2021cardiovascular}
Al-Aly, Z., Xie, Y., Bowe, B.: High-dimensional characterization of post-acute sequelae of COVID-19. Nature \textbf{594}(7862), 259--264 (2021)

\bibitem{pfaff2022identifying}
Pfaff, E.R., Girvin, A.T., Bennett, T.D., Bhatia, A., Brooks, I.M., Deer, R.R., Dekermanjian, J.P., Jolley, S.E., Kahn, M.G., Kostka, K., et al.: Identifying who has Long COVID in the USA: A machine learning approach using N3C data. The Lancet Digital Health \textbf{4}(7), e532--e541 (2022)

\bibitem{brat2020international}
Brat, G.A., Weber, G.M., Gehlenborg, N., Avillach, P., Palmer, N.P., Chiovato, L., Cimino, J., Waitman, L.R., Omenn, G.S., Malovini, A., et al.: International electronic health record-derived COVID-19 clinical course profiles: The 4CE Consortium. NPJ Digital Medicine \textbf{3}(1), 109 (2020)

\end{thebibliography}

\end{document}